\documentclass{article}

\usepackage[nonatbib, preprint]{neurips_2023}

\usepackage[utf8]{inputenc} 
\usepackage[T1]{fontenc}    
\usepackage{hyperref}       
\usepackage{url}            
\usepackage{booktabs}       
\usepackage{amsfonts}       
\usepackage{amssymb}
\usepackage{amsmath}
\usepackage{nicefrac}       
\usepackage{microtype}      
\usepackage[dvipsnames]{xcolor}
\usepackage{graphicx}
\usepackage{lipsum}
\usepackage{caption}
\usepackage{subcaption}
\usepackage{wrapfig}
\usepackage{threeparttable}
\usepackage{csquotes}
\usepackage{bm}
 \usepackage{enumitem}
\newcommand{\loss}{\mathcal{L}}
\newcommand{\bx}{\mathbf{x}}
\newcommand{\by}{\mathbf{y}}
\newcommand{\bz}{\mathbf{z}}
\newcommand{\bp}{\mathbf{p}}
\newcommand{\bw}{\mathbf{w}}
\newcommand{\bh}{\mathbf{h}}
\newcommand{\similar}[2]{\text{sim}(#1,#2)}
\newcommand{\bcdot}{\boldsymbol{\cdot}}
\newcommand{\with}{w\slash\ }
\newcommand{\plus}[1]{\small{\textcolor{ForestGreen}{$\uparrow$#1}}}

\newcommand{\bcheck}{$\bm{\surd}$}
\newcommand{\btimes}{$\bm{\times}$}

\title{Generalized Supervised Contrastive Learning}

\author{%
  Jaewon Kim\\
  Seoul National University Graduate School \\
  \texttt{rlawodnjs017@snu.ac.kr} \\
  \And
  Hyukjong Lee\\
  University of Seoul \\
  \texttt{jj770206@uos.ac.kr} \\
  \And
  Jooyoung Chang \\
  XAIMED Co., Ltd. \\
  \texttt{jchang@ixaimed.com} \\
  \And
  Sang Min Park \\
  Seoul National University Graduate School \\
  \texttt{fmpark1@snu.ac.kr}
}

\begin{document}

\maketitle

\begin{abstract}
With the recent promising results of contrastive learning in the self-supervised learning paradigm, 
supervised contrastive learning has successfully extended these contrastive approaches to supervised contexts, outperforming cross-entropy on various datasets.
However, supervised contrastive learning inherently employs label information in a binary form--either positive or negative--using a one-hot target vector. 
This structure struggles to adapt to methods that exploit label information as a probability distribution, such as CutMix and knowledge distillation.
In this paper, we introduce a generalized supervised contrastive loss, which measures cross-entropy between label similarity and latent similarity. 
This concept enhances the capabilities of supervised contrastive loss by fully utilizing the label distribution and enabling the adaptation of various existing techniques for training modern neural networks.
Leveraging this generalized supervised contrastive loss, we construct a tailored framework: the Generalized Supervised Contrastive Learning (GenSCL). 
Compared to existing contrastive learning frameworks, GenSCL incorporates additional enhancements, including advanced image-based regularization techniques and an arbitrary teacher classifier. 
When applied to ResNet50 with the Momentum Contrast technique, GenSCL achieves a top-1 accuracy of 77.3\% on ImageNet, a 4.1\% relative improvement over traditional supervised contrastive learning. 
Moreover, our method establishes new state-of-the-art accuracies of 98.2\% and 87.0\% on CIFAR10 and CIFAR100 respectively when applied to ResNet50, marking the highest reported figures for this architecture.
\end{abstract}

\section{Introduction}
In the field of computer vision, both generative and discriminative approaches have become dominant in self-supervised visual representation learning. While pixel-level generation in the input space \cite{vae, gan} is computationally expensive and inefficient for representation learning, discriminative approaches, which learn from pseudo labels defined by pretext tasks, have made significant strides in self-supervised learning. Earlier, discriminative approaches relied on heuristic tasks \cite{context_prediction, colorization, jigsaw, rotation}, but the results indicated that heuristics might have limitations in generalization. Recently, contrastive approaches have emerged and demonstrated great promise in addressing the weaknesses of traditional heuristic discriminative approaches, achieving state-of-the-art performances that are even comparable to supervised learning \cite{multiview, moco, mocov2, simclr}. 

The fundamental concept of contrastive learning is the binary classification of positive/negative contrasts with respect to an anchor. In the self-supervised context, where no label information is available, a data augmentation of an anchor is considered as a positive contrast \cite{moco, simclr}. Supervised contrastive learning \cite{supcon} (SupCon) has extended contrastive learning to the supervised context. By leveraging label information, all contrasts from the same class within the minibatch are considered positive contrasts, and contrasts from other classes are negative. The ResNet \cite{resnet} encoder trained on SupCon has outperformed cross-entropy loss on CIFAR \cite{cifar} and ImageNet \cite{imagenet}, demonstrating the significant potential of contrastive learning in the supervised context.

\begin{figure*}[t]
    \centering
    \includegraphics[width=\textwidth]{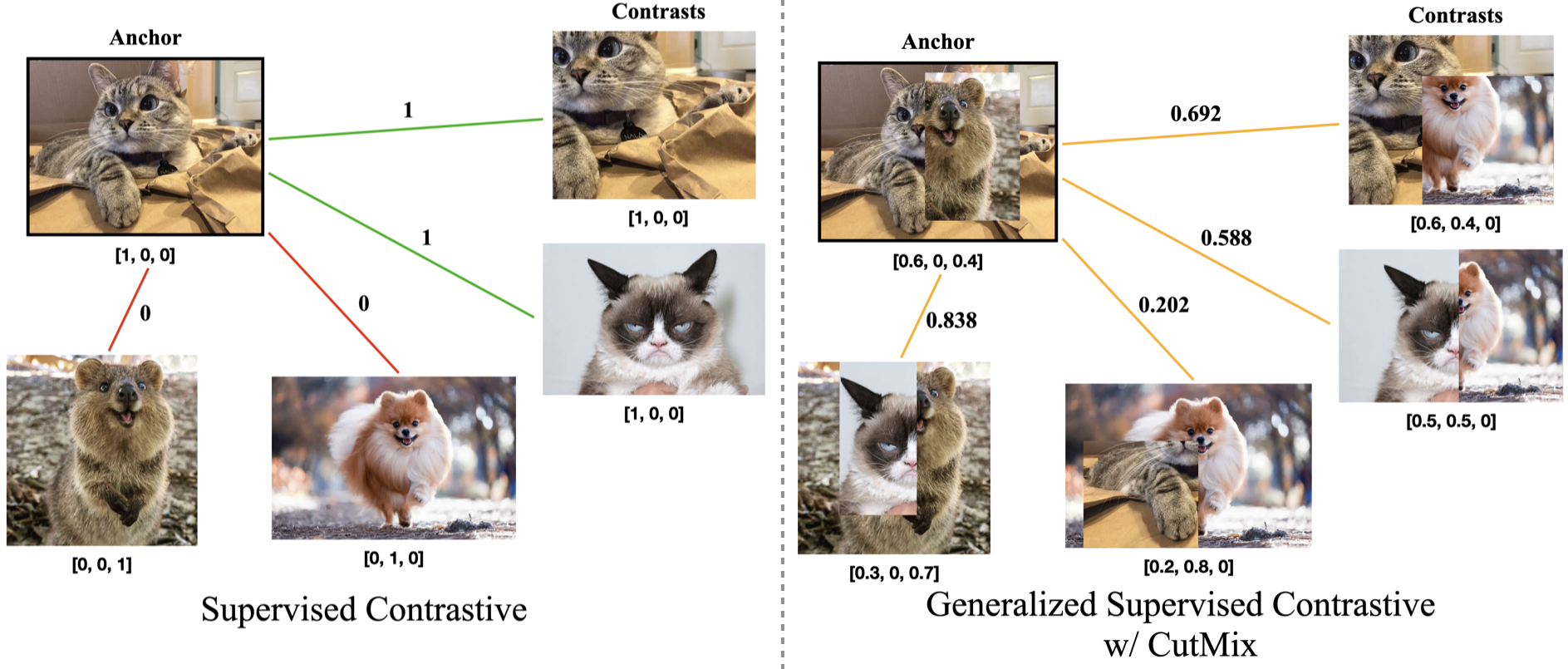}
    \caption{Generalized Supervised \emph{vs.} Supervised Contrastive Losses: The supervised contrastive loss inherently utilizes label information in a binary form (left, Eq. \ref{eq:supcon}), making it incapable of computing the extent of similarity between a mixed anchor and mixed contrasts. In contrast, the generalized supervised contrastive loss measures cross-entropy between label similarity and latent similarity (right, Eq. \ref{eq:gensupcon}), enabling it to perform complex comparisons.}
\label{fig:overview}
\end{figure*}

Given that the overconfidence problem of modern deep neural networks is one of the main obstacles to their generalization \cite{calibration}, a number of studies have proposed regularization methods corresponding to cross-entropy loss. These regularization methods can be practically divided into a) image-based regularization, including data augmentation \cite{autoaugment, randaugment} and mixing techniques \cite{mixup, cutmix}, and b) soft targets, including label smoothing \cite{smoothing, whensmoothing} and knowledge distillation \cite{kd, noisystudent}. Correspondingly, recent work \cite{usi} has shown that a well-organized framework consisting of image-based regularization \cite{randaugment, mixup, cutmix} and knowledge distillation with a proper optimization setting \cite{adamw, onecycle} can train various architectures \cite{resnet, vit, mlpmixer} to top results. However, these regularization methods commonly handle label information as a probability distribution and struggle to adapt directly to supervised contrastive learning, since supervised contrastive loss inherently employs label information in a binary form, as depicted in Figure \ref{fig:overview}(left). Thus, a crucial question arises:

\begin{displayquote}
\emph{How can we fully utilize the label distribution in the contrastive learning paradigm to adapt modern regularization methods?}
\end{displayquote}

In this work, we propose a generalized supervised contrastive loss, which measures cross-entropy between label similarity and latent similarity. This approach builds on the supervised contrastive loss but is capable of leveraging label information as a probability distribution. Intuitively, while supervised contrastive loss and previous self-supervised contrastive loss \cite{moco, simclr} simply divide contrasts into positive and negative by one-hot labels and draw all positive contrasts closer to an anchor evenly, the novelty of our loss is to consider drawing positive contrasts to an anchor delicately according to the label similarity, as shown in Figures \ref{fig:overview}(right) and \ref{fig:kd}. 
Our loss can be seen as a generalization of self-supervised contrastive loss \cite{moco, simclr}, its convex combination \cite{mixco, imix, unmix} and supervised contrastive loss. When utilizing label information in a binary form, the generalized supervised contrastive loss degenerates to conventional contrastive losses. To the best of our knowledge, this is the first contrastive loss to leverage label information in a probability distribution and truly achieves congruence between latent feature space and label space. 
Furthermore, this is the first attempt to adapt knowledge distillation within the contrastive learning paradigm, as shown in Table \ref{tab:loss_comp}.

\begin{table}[h]
    \caption{Comparison of generalized supervised contrastive loss with conventional contrastive losses. Generalized supervised contrastive loss fully leverages label information as a probability distribution. Here, "anch." denotes anchor, "contr." denotes contrasts, and "KD" denotes knowledge distillation.}
    \centering
    \begin{threeparttable}
    \begin{tabular}{ccccc}
    \toprule
     & anch. $\leftrightarrow$ contr. & anch. $\leftrightarrow$ mixed contr. & mixed anch. $\leftrightarrow$ mixed contr. & KD \\
    \midrule
    SupCon & \bcheck & \btimes & \btimes & \btimes \\
    Convex comb.\tnote{$\dagger$} & \bcheck & \bcheck & \btimes & \btimes\\
    GenSCL & \bcheck & \bcheck & \bcheck & \bcheck \\
    \bottomrule
    \end{tabular}
    \begin{tablenotes}
      \item [$\dagger$] Although the convex combination of contrastive loss is proposed in self-supervised context \cite{mixco, imix, unmix}, it can simply be extended to supervised contrastive loss.
    \end{tablenotes}
    \end{threeparttable}
\label{tab:loss_comp}
\end{table}
Leveraging the generalized supervised contrastive loss, we also present a tailored learning framework: the Generalized Supervised Contrastive Learning (GenSCL), which seamlessly adapts mixing techniques \cite{mixup, cutmix} and knowledge distillation \cite{kd}, as shown in Figure \ref{fig:genscl}. This can be considered a contrastive version of the unified scheme for ImageNet (USI) \cite{usi}. The ResNet50 trained on GenSCL consistently outperforms SupCon significantly, as our empirical results show. It achieves a top-1 accuracy of 77.3\% on ImageNet \cite{imagenet}, a 4.1\% relative improvement over supervised contrastive learning with the Momentum Contrast technique \cite{moco} from scratch. Moreover, our method establishes new state-of-the-art accuracies of 98.2\% and 87.0\% on CIFAR10 \cite{cifar} and CIFAR100 \cite{cifar} respectively without external datasets, marking the highest reported figures for this architecture. Our main contributions are summarized below:
\begin{enumerate}[leftmargin=*]
    \item We propose a novel extension to the supervised contrastive loss function that fully utilizes label information in a probability form, thus defining congruence between the latent feature space and label space.
    \item We introduce a tailored contrastive learning framework based on generalized supervised contrastive learning, named GenSCL, which seamlessly adapts mixing techniques and knowledge distillation.
    \item We analytically and empirically demonstrate the effect of mixing techniques during contrastive training. These techniques generate hard positive contrasts to prevent gradient vanishing during training.
    \item We empirically show that the latent features learned by GenSCL are more disentangled.
\end{enumerate}

\section{Related Work}

\paragraph{Contrastive Visual Representation Learning}
Contrastive learning operates on the principle of drawing positive contrasts closer in the representational space while distancing negative contrasts. Traditional self-supervised learning methodologies in the visual domain, such as those detailed in \cite{counting, rr, jigsaw, colorization, rotation}, have depended on pretext tasks for learning representations. Taking cues from noise contrastive estimation \cite{gutmann2010noise, mnih2013learning} and N-pair losses \cite{sohn2016improved}, recent advancements have facilitated the development of discriminative methods within the contrastive learning paradigm, consistently yielding state-of-the-art results \cite{wu2018unsupervised, henaff2020data, multiview, simclr, moco, mocov2}. These methodologies distinguish between positive and negative pairs relative to an ``anchor'' within a given minibatch. In works such as \cite{simclr, moco, mocov2}, only images derived from the same anchor are considered positive. In addition, \cite{supcon} expands the concept of positive pairs by incorporating labels in a supervised manner.

\paragraph{Mixing Strategies for Contrastive Learning}
Data augmentation strategies play an essential role in enhancing the generalization capabilities of deep neural networks, with robust augmentations like \cite{autoaugment, randaugment} leading to improved performance. Specifically, image-based regularizations \cite{cutout, mixup, cutmix} have resulted in a significant leap in performance, establishing them as a prevalent tool for achieving state-of-the-art results in cross-entropy optimization.  The idea of adapting mixup \cite{mixup} to self-supervised learning has been investigated in \cite{verma, shen, zhoug}. However, these methods are limited, as they only function as augmentations. Recently, \cite{mixco, imix, unmix} proposed a convex combination of contrastive loss to compare a mixed anchor with contrasts, which has led to performance enhancements in self-supervised learning. Furthermore, \cite{mixcota, mixcoti} demonstrate that this approach is also effective in other domains.

\paragraph{Knowledge Distillation for Contrastive Learning}
Knowledge distillation \cite{kd} is a concept wherein a ``student'' network is guided by a pre-trained ``teacher'' network. In addition to cross-entropy, the Kullback–Leibler divergence between the predictions of the student and teacher networks can enhance the training of the student model. A recent study \cite{usi} demonstrated that training strategies that incorporate modern techniques and knowledge distillation can consistently achieve top-tier results on cross-entropy optimization. In the realm of contrastive learning, the concept of contrastive representation distillation was introduced by \cite{conrep}. In this approach, the student model learns representations by optimizing the mutual information between the representations of the teacher and student at the penultimate layer.

\section{Method}
\subsection{Generalized Supervised Contrastive Learning Framework}
In this section, we present our proposed Generalized Supervised Contrastive Learning (GenSCL) framework. Our method is designed to adapt seamlessly to mixing techniques and knowledge distillation. GenSCL extends the traditional Supervised Contrastive Learning by leveraging a generalized supervised contrastive loss. This approach can be considered as a contrastive modification of USI \cite{usi}.
\begin{figure*}[t]
    \centering
    \includegraphics[width=\textwidth]{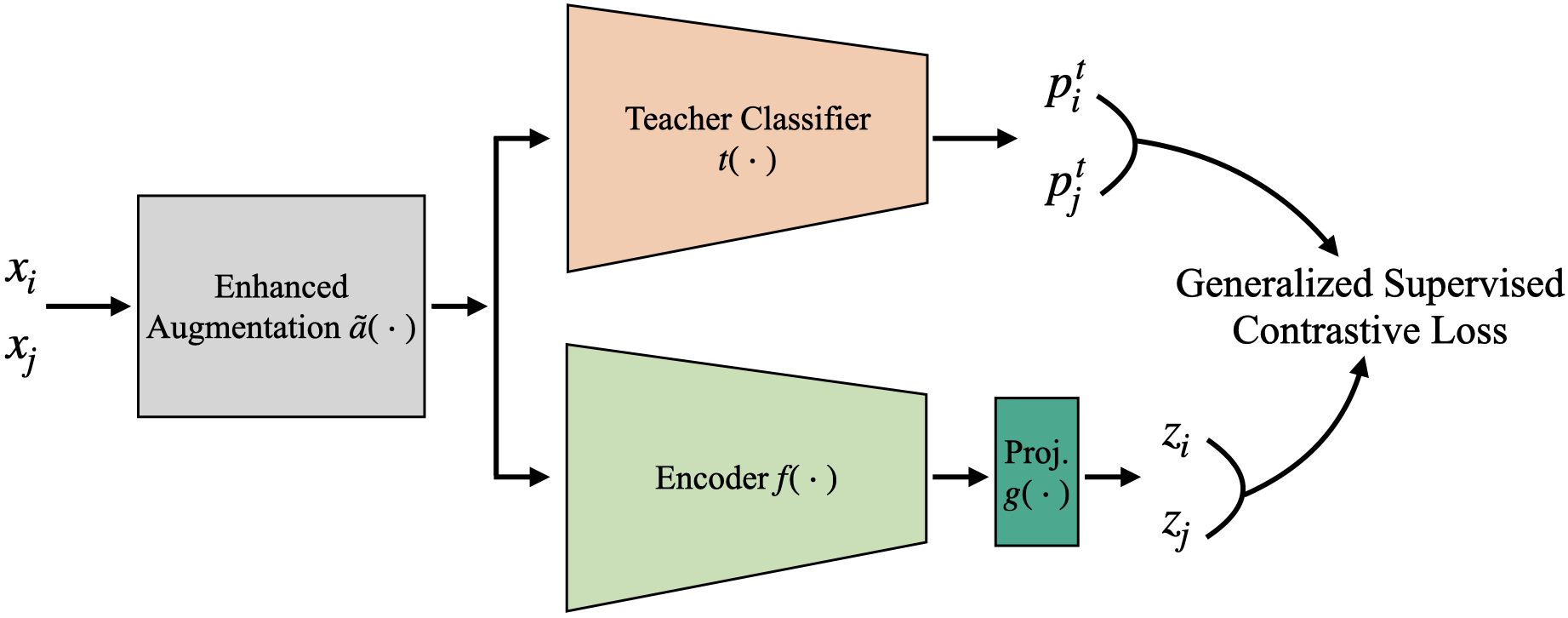}
    \caption{Generalized supervised contrastive learning (GenSCL) framework.}
    \label{fig:genscl}
\end{figure*}
The GenSCL framework, as illustrated in Figure \ref{fig:genscl}, includes the following key components (the notations are referred from \cite{simclr}):

\begin{itemize}[leftmargin=*]
    \item \emph{Enhanced Data Augmentation $\tilde{a}(\cdot)$}: We extend conventional data augmentation $a(\cdot)$, which generates two disparate views of the data \cite{simclr, moco}, by applying additional mixing techniques (such as MixUp \cite{mixup} and CutMix \cite{cutmix}), denoted as $m(\cdot)$. For each data/label pair $(\bx, \by)$ where $\bx \in \mathbb{R}^D$ and $\by \in \mathbb{R}^C$ (where $D$ denotes the data dimension and $C$ represents the number of classes), we generate two unique mixed views of the data and label (i.e., $\tilde{\bx}=\tilde{a}(\bx)=m(a(\bx))$ and $\tilde{\by}=m(\by)$), rendering $\tilde{\by}$ as a probability distribution. In previous contrastive learning frameworks without the application of mixing techniques $m(\cdot)$ \cite{simclr, moco, supcon}, $\tilde{\bx}=a(\bx)$ and $\tilde{\by}=\by$, with $\tilde{\by}$ represented as a one-hot vector.

    \item \emph{Encoder Network $f(\cdot)$}: This module employs a deep neural network backbone for mapping $\tilde{x}$ to latent features $\bh=f(\tilde{\bx})$ where $\bh \in \mathbb{R}^E$, effectively serving as the student in a knowledge distillation scheme. For all our experiments, we use ResNet50 for $f$, with $E=2048$.

    \item \emph{Projection Network $g(\cdot)$}: This network is a two-layer Multi-Layer Perceptron (MLP) equipped with ReLU and normalization, mapping latent features to the unit hypersphere where the contrastive loss is measured. Here, $\bz=g(\bh) \in \mathbb{R}^P$ with $P=128$. Upon completion of the contrastive learning phase, we discard $g(\cdot)$ and substitute it with a linear classifier.

    \item \emph{Teacher Classifier $t(\cdot)$}: This is an arbitrary pre-trained classifier that outputs the predicted label of the augmented data, denoted as $\bp^t=t(\tilde{\bx}) \in \mathbb{R}^C$.

\end{itemize}

\subsection{Contrastive Loss Functions}
Given this framework, we briefly review the supervised contrastive loss \cite{supcon} and its limitations.
Then we demonstrate how the generalized supervised contrastive loss fully utilizes label information in a probability distribution and analyze the validity of GenSCL under generalized supervised contrastive loss.
For the remainder of this paper, we denote a set of $N$ randomly sampled batch data/label pairs as $\{ \bx_k, \by_k \}_{k=1...N}$ and corresponding mixed multi-viewed batch as $\{ \tilde{\bx}_\ell, \tilde{\by}_\ell\}_{\ell=1...2N}$, where $\tilde{\bx}_{2k}$ and $\tilde{\bx}_{2k-1}$ are two different views of $\bx_{k}$ ($k=1...N$).
Within the multi-viewed batch, let $i \in I \equiv \{ 1...2N \}$ be the index of an \emph{anchor} and $A(i) \equiv I \setminus \{ i \}$ is the set of \emph{contrasts} with respect to an anchor $i$.

\subsubsection{Supervised Contrastive Loss}
We denote the superior supervised contrastive loss \cite{supcon} $\loss^{\text{sup}}_{\text{out}}$ as $\loss^{\text{sup}}$ as in Eq. \ref{eq:supcon}.
For the supervised contrastive loss, the label information is implicitly embedded in $P(i) \equiv \{ p \in A(i): \tilde{\by}_p=\tilde{\by}_i\}$, which means the positive contrasts set $P(i)$ with respect to anchor $i$ can only be selected from binary comparison with one-hot encoded labels.
Thus, when comparing a mixed anchor with mixed contrasts, meaning $\tilde{\by}$ is in the form of a probability distribution, supervised contrastive learning is not applicable as the positive contrasts set $P(i)$ is approximately an empty set.
Equivalently, constructing the positive contrasts set, which reflects the prediction of teacher $\bp^t$ in the form of a probability distribution for knowledge distillation, is challenging.
Thus, neither comparison between mixed anchors and mixed contrasts nor prediction labels of the teacher classifier are applicable to the original supervised contrastive loss.

\begin{equation}
    \label{eq:supcon}
    \loss^{\text{sup}} = \sum\limits_{i \in I} \loss^{\text{sup}}_i \\
    = \sum\limits_{i \in I} \frac{-1}{|P(i)|} \sum\limits_{j \in P(i)} \log \frac{\exp(\bz_i \bcdot \bz_j / \tau)}{\sum\limits_{a \in A(i)} \exp(\bz_i \bcdot \bz_a / \tau)}
\end{equation}
Here, $\bz_\ell=g(f(\tilde{\bx}_\ell)) \in \mathbb{R}^{E}$, the symbol $\bcdot$ denotes the dot product, $\tau \in \mathbb{R}^+$ is a scalar temperature parameter, and $|P(i)|$ is its cardinality.

\subsubsection{Generalized Supervised Contrastive Loss}
To fully utilize the label information in the form of the probability distribution, generalized supervised contrastive aims to minimize cross-entropy between label similarity space $Y(i) \equiv \{ \similar{\boldsymbol{y}_i}{\boldsymbol{y}_\ell} \}_{\ell \in A(i)}$ and latent similarity space $Z(i) \equiv \{ P_{i \ell} \}_{\ell \in A(i)}$ with respect to anchor $i$, where $\similar{\boldsymbol{u}}{\boldsymbol{v}}=\boldsymbol{u}^\intercal \boldsymbol{v} / \Vert \boldsymbol{u} \Vert \Vert \boldsymbol{v} \Vert$ (i.e., cosine similarity of labels) and $P_{ij}= \nicefrac{\exp(\bz_i \bcdot \bz_j / \tau)}{\sum\limits_{a \in A(i)} \exp(\bz_i \bcdot \bz_a / \tau)}$ (i.e., the similarity of projected latent features). Then, the generalized supervised contrastive loss takes the following form:
\begin{equation}
    \label{eq:gensupcon}
    \begin{aligned}
    \loss^{\text{gen}} = \sum\limits_{i \in I} \loss^{\text{gen}}_i 
    &= \sum\limits_{i \in I} \frac{1}{|A(i)|} \text{CE}(Y(i), Z(i)) \\
    &= \sum\limits_{i \in I} \frac{-1}{|A(i)|} \sum\limits_{j \in A(i)} \similar{\by_i}{\by_j} \log \frac{\exp(\bz_i \bcdot \bz_j / \tau)}{\sum\limits_{a \in A(i)} \exp(\bz_i \bcdot \bz_a / \tau)}
    \end{aligned}
\end{equation}
Here $\text{CE}$ denotes cross-entropy loss. As shown in Eq. \ref{eq:gensupcon}, generalized supervised contrastive loss $\loss^{\text{gen}}$ no longer depends on the positive contrasts set $P(i)$ which enforces the use of one-hot encoded labels.
Intuitively, the generalized supervised contrastive loss considers drawing positive contrasts to an anchor delicately according to label similarity, instead of pulling all positive contrasts evenly as in previous contrastive losses \cite{moco, simclr, supcon}.

\subsubsection{Intrinsic Hard Positive/Negative Mining Property}
\label{sec:mining}
The projection network performs normalization on its outputs. Let's consider $\bw_i$ to represent the output of the projection network before normalization, which implies $\bz_i=\bw_i / ||\bw_i||$. As highlighted in \cite{supcon}, normalizing projected features within projection networks inherently enables hard positive/negative mining.

\begin{equation}
    \label{eq:hardmining}
    \frac{\partial \loss^{\text{Gen}}_{i}}{\partial \bw_i} = 
    \frac{1}{\tau ||\bw_i||}
            \left[
                \sum\limits_{j \in A(i)}
                \left(
                    \bz_j - (\bz_i \bcdot \bz_j) \bcdot \bz_i
                \right)
                \left\{
                    \frac{\sum\limits_{a \in A(i)} \similar{\by_i}{\by_a}}{|A(i)|}P_{ij}
                    -
                    \frac{\similar{\by_i}{\by_j)}}{|A(i)|}
                \right\}
            \right]
\end{equation}
As shown in Eq. \ref{eq:hardmining} (a full derivation is provided in the Supplementary material), the intrinsic hard positive/negative mining property still holds for the generalized supervised contrastive loss, since the gradient contribution for $\loss^{\text{Gen}}_i$ with respect to the embedding $\bw_i$ depends on $|| \bz_j - (\bz_i \bcdot \bz_j) \bcdot \bz_i ||$. 
For easy positive/negative (i.e., $\bz_i \bcdot \bz_j = \pm 1$), the gradient contribution converges to zero as shown below:
\begin{equation}
    \label{eq:easygradient}
    || \bz_j - (\bz_i \bcdot \bz_j) \bcdot \bz_i || = \sqrt{1 - (\bz_i \bcdot \bz_j)^2} \approx 0
\end{equation}
which is identical to the finding in supervised contrastive learning.
As contrastive training progress, it is reasonable to assume that the model will learn proper feature representations and the comparison between an anchor and the contrasts will become easier, which means the gradient contribution will ``vanish''. 
However, in GenSCL, since both $\bz_i$ and $\bz_j$ are extracted from mixed data, the gradient vanishing problem could be prevented by hard comparisons (i.e., $\bz_i \bcdot \bz_j \approx 0$), aligning with the idea that heavy data augmentations are crucial for efficient representation learning as revealed in recent studies \cite{asano2019critical, gontijo2020affinity, wang2020understanding, tian2020makes}.

\subsection{Knowledge Distillation}
We first briefly review conventional knowledge distillation \cite{kd} applied for cross-entropy training. 
Then, we will discuss how to seamlessly adapt knowledge distillation to the contrastive learning paradigm in GenSCL.

\subsubsection{Knowledge Distillation in cross-entropy training}
Given an image, a classifier outputs a logit vector $\bp=\{ p_i \}_{i=1...C}$, where $C$ is the number of classes. 
With label-smoothing regularization \cite{smoothing}, the softened prediction vector is denoted as $\boldsymbol{p}(\tilde{\tau})=\{ \tilde{p}_i (\tilde{\tau}) \}_{i=1...C}$, where 
\begin{equation}
    \tilde{p}_i(\tilde{\tau})=\frac{\exp(p_i/\tilde{\tau})}{\sum^C_{j=1}\exp(p_j/\tilde{\tau})}
\end{equation}
Here, $\tilde{\tau} \in \mathbb{R}^+$ is a scalar temperature parameter for the softmax activation function. 
For convenience, we set $\tilde{\tau}=1$ in this paper, and thus, $\boldsymbol{p}(\tilde{\tau})$ is simply denoted as $\boldsymbol{p}$.
In conventional knowledge distillation, the objective function for training the student classifier is a combination of cross-entropy between predictions of the student classifier ($\bp^s \in \mathbb{R}^C$) and ground-truth labels ($\by \in \mathbb{R}^C$), and the Kullback-Leibler (KL) divergence between the predictions of the student and teacher classifier ($\bp^t \in \mathbb{R}^C$),
\begin{equation}
    \label{eq:kd}
    \loss = \loss_{\text{CE}}(\bp^s, \by) + \alpha_{\text{kd}} \loss_{\text{KL}}(\bp^s, \bp^t)
\end{equation}
where $\alpha_{\text{kd}}$ is a hyper-parameter for adjusting the weight of knowledge distillation loss. 

\subsubsection{Knowledge Distillation in GenSCL}
\begin{figure*}[t]
    \centering
    \includegraphics[width=0.5\linewidth]{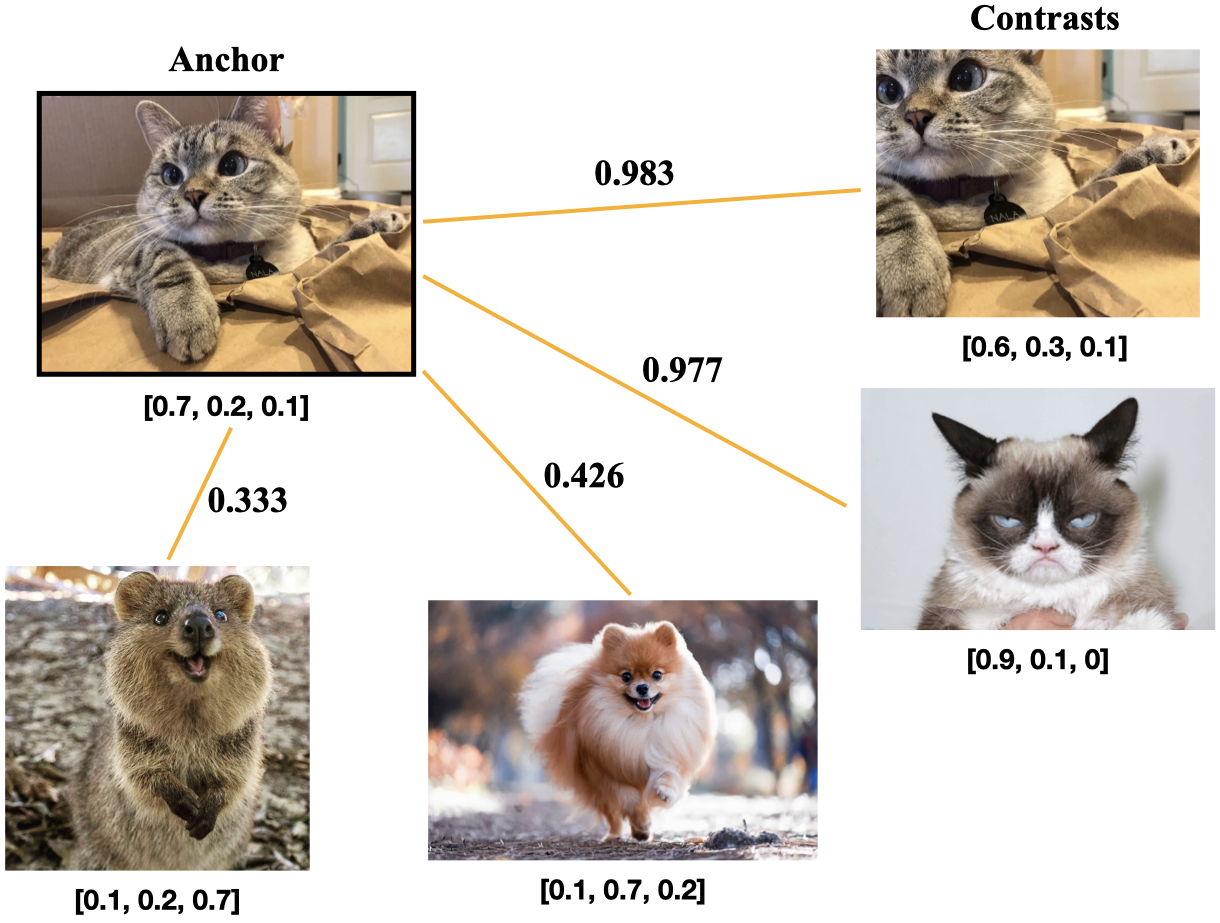}
    \caption {Generalized supervised contrastive learning with knowledge distillation. It achieves the congruence between teacher prediction space and latent space.}
\label{fig:kd}
\end{figure*}
Leveraged by the generalized supervised contrastive loss, as shown in Figure \ref{fig:kd}, the basic idea of applying knowledge distillation in the contrastive learning paradigm is to minimize the weighted cross-entropy between the teacher prediction similarity space $P^t(i) \equiv \{\similar{\boldsymbol{p}^t_i}{\boldsymbol{p}^t_\ell}\}_{\ell \in A(i)}$ and the latent similarity space $Z(i) \equiv \{ P_{i \ell} \}_{\ell \in A(i)}$ with respect to anchor $i$. 
Then, similar to the original knowledge distillation in cross-entropy, the knowledge distillation in GenSCL takes the following form:
\begin{equation}
    \label{eq:gensupcon_kd}
    \begin{aligned}
        \loss^{\text{kd-gen}} &= \sum\limits_{i \in I} \frac{1}{|A(i)|} \left\{ \text{CE}(Y(i), Z(i)) + \alpha_{\text{kd}} \text{CE}(P^t(i), Z(i)) \right\} \\
        &= \sum\limits_{i \in I} \frac{-1}{|A(i)|} \sum\limits_{j \in A(i)} \left\{ \left( \similar{\by_i}{\by_j} + \alpha_{\text{kd}} \similar{\bp^t_i}{\bp^t_j} \right) \log \frac{\exp(\bz_i \bcdot \bz_j / \tau)}{\sum\limits_{a \in A(i)} \exp(\bz_i \bcdot \bz_a / \tau)} \right\}
    \end{aligned}
\end{equation}

\section{Experiments}
In this section, we first evaluate our generalized supervised contrastive learning framework by measuring classification accuracy on common benchmarks - CIFAR10/100 \cite{cifar} and ImageNet \cite{imagenet} (Section \ref{sec:clsacc}).
For a fair comparison with the original supervised contrastive learning, for CIFAR10/100, we use ResNet50 \cite{resnet} as the encoder architecture in all experiments and maintain the same hyper-parameter settings as in \cite{supcon}.
For ImageNet, we implement both supervised contrastive learning and generalized supervised contrastive learning with the Momentum Contrastive technique \cite{moco} from scratch.
We demonstrate that the adaptation of mixing techniques and knowledge distillation techniques by GenSCL can significantly improve classification accuracies.
For the selection of teacher models, to ensure a fair comparison, we leverage CIFAR-only trained PyramidNet272 \cite{pyramid} provided by \cite{cifar_teacher} (with respective accuracies of 98.7\% and 89.0\% on CIFAR10 and CIFAR100) and ImageNet1K-only trained ConvNeXt-large \cite{imagenet_teacher} provided by \cite{timm} (with an accuracy of 84.8\%) as pre-trained teacher classifiers.
Next, we empirically show that mixing techniques can prevent supervised contrastive learning from experiencing gradient vanishing caused by easy comparisons in Section \ref{sec:effect}.
We then demonstrate that the encoder in GenSCL learns more disentangled representations through feature visualization in Section \ref{sec:visualization}.
Lastly, we provide the training details in Section \ref{sec:details}.

\subsection{Classification Accuracy}
\label{sec:clsacc}
\begin{table}[ht]
  \caption{Top-1 classification accuracy (\%) on ResNet-50 for various datasets.}
  \begin{threeparttable}
  \centering
  \begin{tabular}{cccccc}
    \toprule
     & & & \multicolumn{3}{c}{GenSCL} \\
    \cmidrule(r){4-6}
    Dataset & Self-Sup Con \cite{simclr,mocov2} & SupCon & \with CutMix & \with KD & \with CutMix \& KD \\ 
    \midrule
    CIFAR10 & 93.6 & 96.0 & 97.1 \plus{1.1} & 97.7 \plus{1.7}& \textbf{98.2 \plus{2.2}} \\
    CIFAR100 & 70.7 & 76.5 & 81.7 \plus{5.2} & 85.3 \plus{8.8} & \textbf{87.0 \plus{10.5}} \\
    ImageNet & 67.5 & 73.2\tnote{$\dagger$} & 76.1 \plus{2.9} & 75.4 \plus{2.2} & \textbf{77.3 \plus{4.1}} \\
    \bottomrule
  \end{tabular}
  \begin{tablenotes}
      \item [$\dagger$] denotes the supervised contrastive learning with Momentum Contrast technique from scratch.
  \end{tablenotes}
  \end{threeparttable}  
  \label{tab:acc}
\end{table}
Table \ref{tab:acc} demonstrates that generalized supervised contrastive learning, when adapted with CutMix \cite{cutmix} and knowledge distillation \cite{kd}, can significantly outperform supervised contrastive learning.
Notably, our method establishes new state-of-the-art accuracies of 98.2\% and 87.0\% on CIFAR10 and CIFAR100, respectively, when applied to ResNet50, marking the highest reported figures for this architecture.
On ImageNet, even though the performance of supervised contrastive learning with Momentum Contrastive techniques is lower than originally reported in \cite{supcon}, GenSCL still achieves a top-1 accuracy of 77.3\%, a 4.1\% relative improvement in the same hyper-parameter setting.

\begin{table}[ht]
  \caption{Top-1 classification accuracy (\%) on various datasets for different image-based regularization methods with ResNet-50.}
  \centering
  \begin{tabular}{cccc}
    \toprule
     & & \multicolumn{2}{c}{GenSCL} \\
    \cmidrule(r){3-4}
    Dataset & SupCon & \with MixUp & \with CutMix \\
    \midrule
    CIFAR10 & 96.0 & 96.7 \plus{0.7} & \textbf{97.1 \plus{1.1}} \\
    CIFAR100 & 76.5 & 80.6 \plus{4.1} & \textbf{81.7 \plus{5.2}} \\
    ImageNet & 73.2 & 74.8 \plus{1.6} & \textbf{76.1 \plus{2.9}} \\
    \bottomrule
  \end{tabular}
  \label{tab:mix}
\end{table}
Table \ref{tab:mix} indicates that both MixUp \cite{mixup} and CutMix \cite{cutmix} lead to significant increases in accuracy when leveraging the generalized supervised contrastive loss. This table also empirically demonstrates that CutMix can result in more substantial improvements in accuracy within the contrastive learning paradigm.

\begin{table}[ht]
  \caption{Top-1 classification accuracy (\%) on various datasets for different knowledge distillation relative weights $\alpha_\text{kd}$ with ResNet-50.}
  \centering
  \begin{threeparttable}
  \begin{tabular}{ccccccc}
    \toprule
     & \multicolumn{2}{c}{CIFAR10} & \multicolumn{2}{c}{CIFAR100} & \multicolumn{2}{c}{ImageNet} \\
     \cmidrule(r){2-3} \cmidrule(r){4-5} \cmidrule(r){6-7}
     $\alpha_{\text{kd}}$ & GenSCL & \with CutMix & GenSCL & \with CutMix & GenSCL & \with CutMix \\
     \midrule
     0 & 96.0\tnote{$\dagger$} & 97.1 \plus{1.1} & 76.5\tnote{$\dagger$} & 81.7 \plus{5.2} & 73.2\tnote{$\dagger$} & 76.1 \plus{2.9} \\
     1 & 97.0 \plus{1.0} & 98.1 \plus{2.1} & 82.6 \plus{6.1} & 86.1 \plus{9.6} & 74.6 \plus{1.4} & 76.9 \plus{3.7} \\
     5 & \textbf{97.7 \plus{1.7}} & \textbf{98.2 \plus{2.2}} & \textbf{85.3 \plus{8.8}} & \textbf{87.0 \plus{10.5}} & 75.1 \plus{1.9} & 77.2 \plus{4.0} \\
     $\infty$\tnote{$\ddagger$} & 97.5 \plus{1.5} & 98.1 \plus{2.1} & 84.7 \plus{8.2} & 86.9 \plus{10.4} & \textbf{75.4 \plus{2.2}} & \textbf{77.3 \plus{4.1}} \\
     \bottomrule
  \end{tabular}
  \begin{tablenotes}
      \item [$\dagger$] degenerates to supervised contrastive learning.
      \item [$\ddagger$] denotes that training relies only on the teacher's predictions.
  \end{tablenotes}
  \end{threeparttable}
  \label{tab:kd}
\end{table}
Table \ref{tab:kd} exhibits the effectiveness of our design of knowledge distillation in GenSCL, as shown in equation \ref{eq:gensupcon_kd}.
The results indicate that combining knowledge distillation with mixing techniques brings further benefits.

\subsection{Effect of Image-based Regularizations on Contrastive Learning}
\label{sec:effect}
\begin{figure*}[t]
    \centering
    \includegraphics[width=\textwidth]{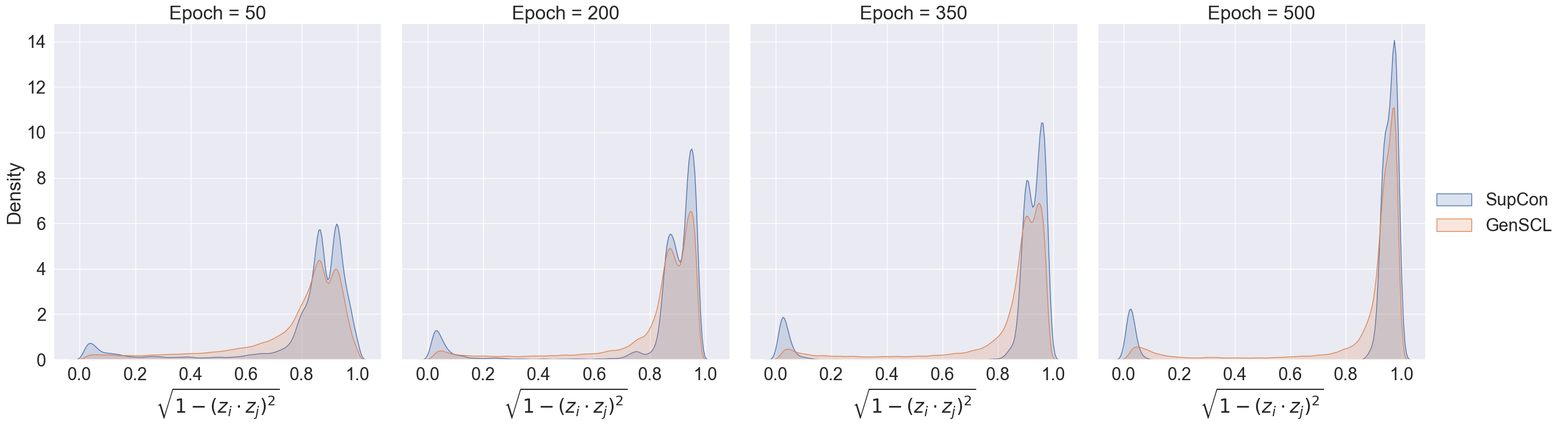}
    \caption{The changes of gradient contribution during training on CIFAR10.}
    \label{fig:contribution}
\end{figure*}
In Section \ref{sec:mining}, we analytically demonstrated that generalized supervised contrastive loss \ref{eq:gensupcon} not only maintains the hard positive/negative mining property but also mitigates the gradient vanishing problem due to our mixed input data by $\bz_i \bcdot \bz_j$ preventing convergence to $= \pm 1$ (i.e., serving as a gradient contribution). Here, we empirically illustrate how $\bz_i \bcdot \bz_j$ operates during the training phase.

In Figure \ref{fig:contribution}, we calculate $\bz_i \bcdot \bz_j$ during training on CIFAR10. As the encoders are trained, $\bz_i \bcdot \bz_j$ for both SupCon \cite{supcon} and GenSCL progressively approach 1. This suggests that as training proceeds, the gradient contribution becomes $\sqrt{1 - (\bz_i \bcdot \bz_j)^2} \approx 0$ thus, consequently causing Eq. \ref{eq:easygradient} to approach 0.

However, when compared to SupCon, GenSCL exhibits a smoother transition toward these extreme values. This behavior suggests that GenSCL maintains a more steady learning trajectory, ensuring that it continually identifies and learns from hard positive/negative contrasts. This intrinsic hard positive/negative mining capability of GenSCL contributes to its superior performance in learning effective representations.

\subsection{Visualization of Learned Representations}
\label{sec:visualization}
\begin{figure}[t]
     \centering
     \begin{subfigure}[b]{0.8\textwidth}
         \centering
         \includegraphics[width=\textwidth]{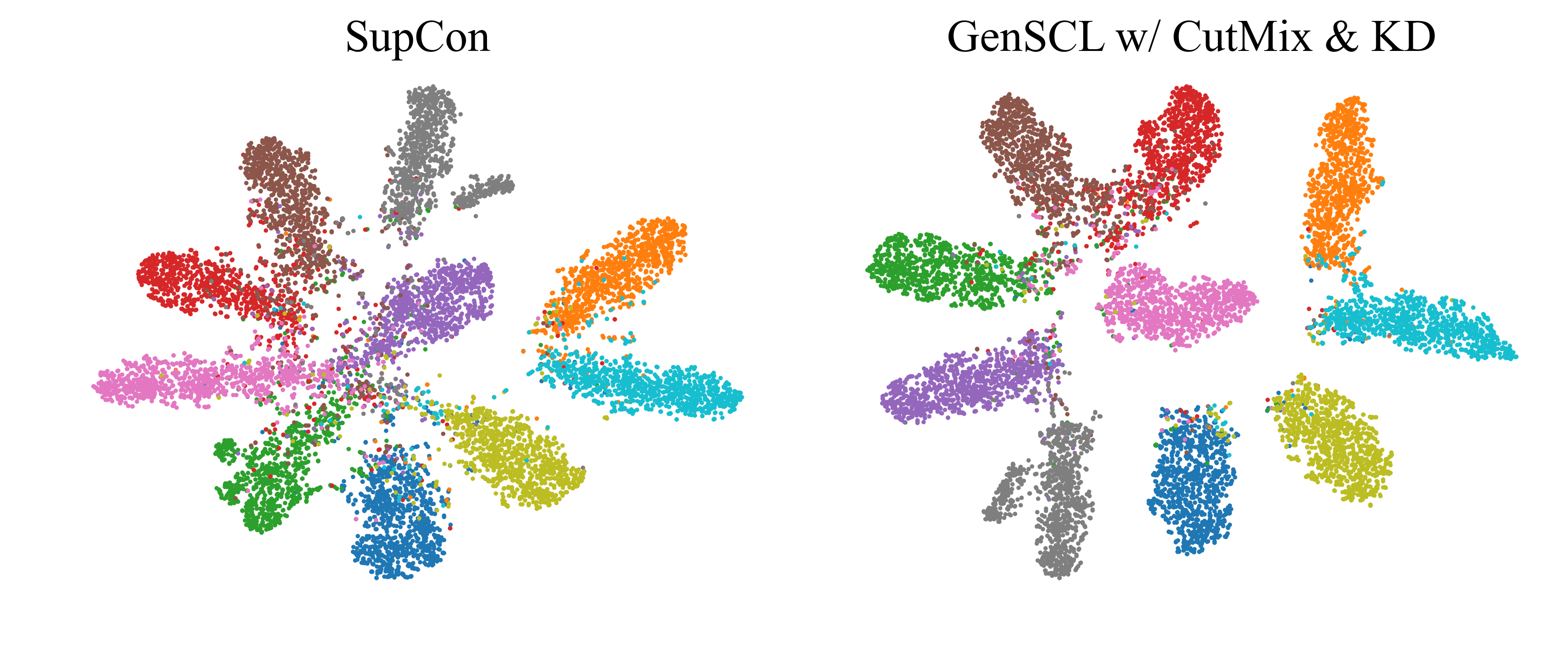}
         \caption{CIFAR10}
         \label{fig:cifar10}
     \end{subfigure}
     \begin{subfigure}[b]{0.8\textwidth}
         \centering
         \includegraphics[width=\textwidth]{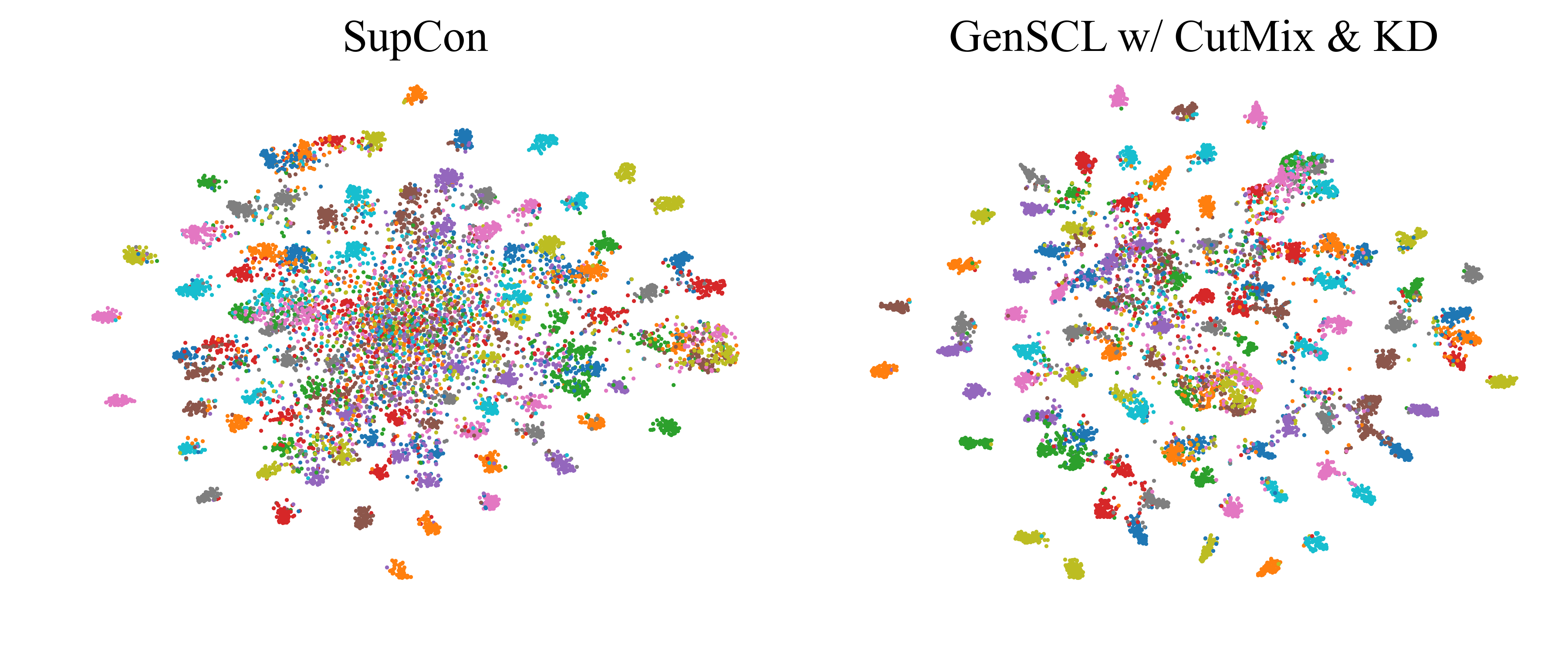}
         \caption{CIFAR100}
         \label{fig:cifar100}
     \end{subfigure}
    \caption{t-SNE visualization of learned representations. Classes are indicated by colors.}
    \label{fig:three graphs}
\end{figure}

t-SNE \cite{tsne}, a powerful non-linear dimensionality reduction and data visualization technique, is utilized to compare the representations between our GenSCL (incorporating CutMix and KD), and SupCon \cite{supcon}. This comparison is conducted on both CIFAR10 and CIFAR100 \cite{cifar}. The resulting t-SNE visualizations offer compelling evidence that the ResNet encoder trained on GenSCL has learned a more disentangled representation. For both datasets, the representation vectors from the same classes cluster together tightly and are well separated from those of other classes.

\subsection{Experimental Setup and Training Details}
\label{sec:details}
In all our experiments, we trained ResNet50 \cite{resnet} as the encoder architecture across all experiments. We trained all models using Stochastic Gradient Descent (SGD) with momentum set at 0.9 and weight decay of 0.0001. For the CutMix \cite{cutmix} in GenSCL, we consistently set the alpha and beta parameters to $1$ for the beta distribution.
In the case of ImageNet \cite{imagenet} training, we utilized the PyTorch distributed library \cite{ddp} across 8 A100-GPUs, training the model for 200 epochs with a batch size of 128. The training process involved the MoCo /\cite{moco} trick. We set the  learning rate at 0.03 and incorporated a warm-up phase during the first 10 epochs, followed by a cosine annealing scheduler for the remaining epochs. Furthermore, we adjusted the temperature parameter of the softmax function to 0.07.
When training on CIFAR10 and CIFAR100 \cite{cifar}, we altered the temperature parameter to 0.1 and trained the model for 500 epochs with a learning rate of 0.5.

For the linear evaluation stage, we implemented a training regimen of 100 epochs for the classifier. This process used SGD as the optimizer, a CrossEntropy loss function, and a batch size of 32. We evaluated the performance of our encoder using top-1 accuracy.
For all other details not explicitly specified, we followed \cite{mocov2}.

\medskip
{
\bibliographystyle{ieee_fullname}
\bibliography{genscl.bib}
}


\end{document}